\documentclass[11pt]{article}
\usepackage{acl}

\usepackage{times}
\usepackage{latexsym}
\usepackage[T1]{fontenc}
\usepackage[utf8]{inputenc}
\usepackage{microtype}

\usepackage{multirow}
\usepackage{amsmath}
\usepackage{amssymb}
\usepackage{booktabs}
\usepackage{graphicx}
\usepackage{hyperref}
\usepackage{url}
\usepackage{xcolor}
\usepackage{xspace}
\usepackage{float}
\usepackage{comment}
\usepackage{algorithm}
\usepackage{algpseudocode}
\usepackage{booktabs}

\newcommand{\sysname}{\textsc{Scale}\xspace}

\usepackage{placeins}

\title{\sysname: Real-Time Latent Communication for Cyber-Physical Agents}

\author{
Shanghao Shi$^1$, Jian Xiao$^1$, Ching-Hsiang Chan$^1$, Sanjoy Baruah$^1$, Sastry Kompella$^2$,\\ \textbf{Yevgeniy Vorobeychik}$^1$, \textbf{William Yeoh}$^1$, \textbf{Ning Zhang}$^1$ \\
$^1$Washington University in St.\ Louis, $^2$Nexcepta Inc.
}

\begin{document}

\maketitle

\begin{abstract}
Latent communication enables agents to collaborate efficiently by directly exchanging intermediate model representations rather than raw sensor data or text messages. Existing approaches, however, are primarily designed for LLM agents and do not consider the bandwidth constraints and heterogeneous task deadlines faced by cyber-physical agents. In this work, we propose \sysname, a real-time latent communication method tailored for cyber-physical agents. Motivated by the spatio-temporal continuity of the physical world, \sysname matches tokens across consecutive latent messages using similarity scores to identify reusable information and avoid redundant transmission. It then profiles the semantic gain and communication cost of transmitting different amounts of latent information. Based on these profiles, \sysname employs a real-time scheduler that jointly determines which pending messages to transmit and how many latent tokens to send, subject to available bandwidth and task deadlines. We evaluate \sysname across three cyber-physical benchmarks and a physical spider robot case study under diverse network conditions and heterogeneous task deadlines. Our results show that \sysname significantly improves both task performance and on-time delivery in bandwidth-constrained networks. Our code is available at: \url{https://anonymous.4open.science/r/SCALE-4E80/}.
\end{abstract}


\section{Introduction}

Communication is fundamental to cyber-physical agents. Individual agents often possess only partial observations or complementary capabilities, and therefore must exchange information with one another to accomplish diverse cyber-physical tasks~\citep{Patel_2021_ICCV, huwhere2comm}. However, transmitting raw sensor observations incurs substantial communication overhead, while natural-language communication introduces additional computational overhead from generating text at the sender and interpreting it at the receiver. These limitations make conventional communication inefficient for real-time multi-agent systems.

\paragraph{Existing Literature and the Gap.} Latent communication has emerged as a potential ``AI language'', allowing agents to directly exchange hidden states or intermediate representations to improve efficiency~\citep{du2026interlat, zou2026latentmas, shi2026selectivekvcomm, ye2025crosscontextkvcomm}. However, existing methods are primarily designed for LLM-based agents and focus on reducing model-side computation, with limited consideration of communication bandwidth and task deadlines.  Meanwhile, communication-efficient collaborative-perception methods consider bandwidth and delay~\citep{liu2020when2com, huwhere2comm, yang2023how2comm, hu2024communication}, but primarily focus on task-specific perception rather than multiple concurrent CPS tasks sharing a bandwidth-constrained link. Thus, latent communication for CPS agents under practical network constraints and heterogeneous task deadlines remains largely underexplored.

\paragraph{Our Work.} We propose \sysname, a real-time latent communication method tailored for cyber-physical agents. \sysname exploits the spatiotemporal continuity of the physical world to reuse latent information already available at the receiver, reducing redundant transmission. It employs a real-time scheduler that jointly determines which messages to transmit and how many latent tokens to send based on semantic gain, available bandwidth, and task deadlines, thereby improving both task utility and message on-time delivery. In addition, \sysname is lightweight and operates on latent tokens produced by existing modality encoders without modifying the underlying models, supporting different sensory modalities and downstream CPS tasks.

Designing \sysname addresses three technical challenges. \textbf{First, spatiotemporal variation makes it difficult to determine what information has actually changed.} 
Physical observations often evolve continuously, but object, agent, or viewpoint motion can make fixed-position latent comparisons overestimate motion for semantic change. \sysname therefore performs \emph{semantic change evaluation} by matching current latent tokens against the receiver's acknowledged latent state and measuring change relative to their best correspondences. 
\textbf{Second, the amount of latent information to transmit must adapt to each message.} 
The semantic gain from transmitting additional latent tokens varies across messages, making a fixed transmission size inefficient. \sysname therefore performs \emph{utility--cost profiling} for each message, ranking latent tokens by semantic change and characterizing the cumulative semantic gain and communication cost at different transmission sizes to guide real-time scheduling.
\textbf{Third, limited and time-varying bandwidth must be allocated across messages with heterogeneous deadlines.} 
Allocating more latent tokens to one message increases its semantic gain but may delay other messages beyond their deadlines. \sysname therefore employs a \emph{real-time scheduler} that greedily allocates latent tokens based on marginal semantic gain per communication cost, while maintaining deadline feasibility for all selected messages under earliest-deadline-first (EDF) ordering.

\begin{figure}[t]
    \centering
    \includegraphics[width=0.5\textwidth]{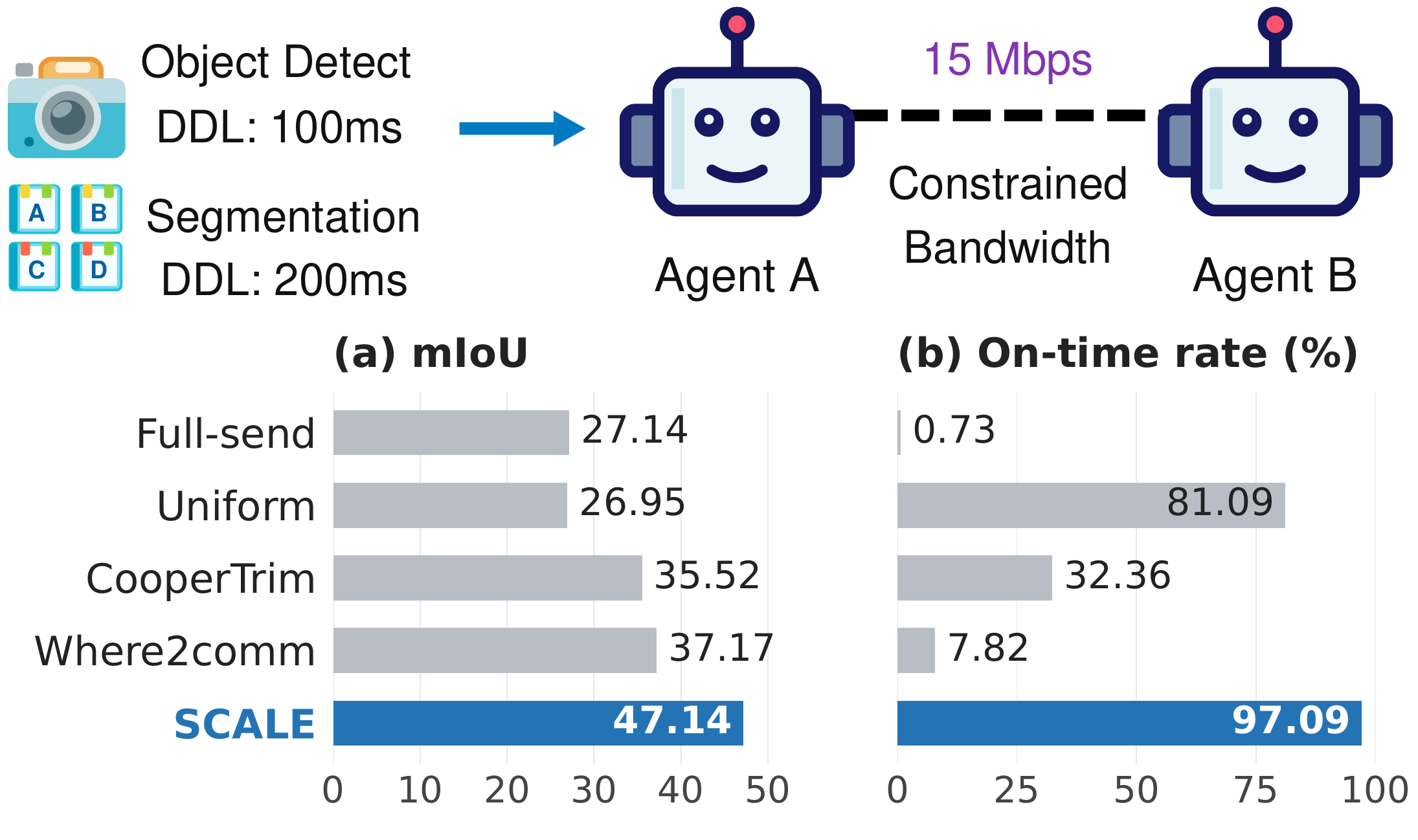}
    \caption{Comparison between \sysname and other latent communication baselines on the OPV2V benchmark at 15 Mbps. \sysname achieves the best mIoU scores and message on-time rates.}
    \label{fig:teaser} 
\end{figure}

We implement and evaluate \sysname on CooperScene \cite{wu2026cooperscene}, OPV2V \cite{xu2022opv2v}, and LIBERO \cite{liu2023libero}, covering autonomous driving and robotic manipulation benchmarks. We further conduct a physical spider robot case study. Our evaluation considers constrained bandwidth, time-varying link capacity, and background traffic, with heterogeneous task deadlines. We compare \sysname against full latent transmission, uniform compression, and four state-of-the-art methods \cite{xu2025cosdh, xu2022v2x, huwhere2comm, mukhopadhyay2026coopertrim}. Our results show that \sysname achieves the highest task performance in almost all settings while significantly improving on-time delivery rates (see Fig.~\ref{fig:teaser} for an example).

We make the following contributions:

\begin{itemize}
\item We identify and formulate the problem of real-time latent communication for CPS agents, where latent messages from multiple downstream tasks with heterogeneous deadlines share a bandwidth-constrained communication link, a setting not adequately addressed by existing methods.
\item We propose \sysname, a real-time latent communication method that combines spatiotemporal latent reuse, utility--cost profiling, and deadline-aware scheduling to improve both task utility and message on-time delivery under bandwidth constraints.
\item We implement and evaluate \sysname across three CPS benchmarks and a physical spider robot case study under diverse network conditions, demonstrating its effectiveness in both benchmark evaluations and a physical robotic system.
\end{itemize}

\section{Related Work}

\paragraph{LLM-based latent communication.}
Recent LLM-based latent communication methods enable multi-agent systems to exchange latent representations, such as hidden states or KV caches, to reduce model inference and computation overhead~\citep{du2026interlat, zou2026latentmas, ye2025crosscontextkvcomm, shi2026selectivekvcomm, fu2026cache, chen2026condenseflow}. These approaches primarily focus on efficient latent interaction and reducing model-side computation, with limited consideration of the underlying communication network. \sysname instead studies latent communication over resource-constrained networks, a practical setting for cyber-physical agents.

\paragraph{Communication-efficient cyber-physical systems.}
Communication-efficient collaborative perception has been extensively studied in cyber-physical systems, where agents exchange intermediate features to improve shared perception~\citep{wang2020v2vnet, li2021disconet}. Existing methods explore when, where, and how agents should communicate through communication triggering~\citep{liu2020when2com}, spatial feature selection~\citep{huwhere2comm}, and feature sparsification~\citep{yang2023how2comm, hu2024communication}. Other work improves robustness to communication delay and asynchrony~\citep{lei2022latency, wei2023asynchrony} or exploits temporal redundancy to reduce repeated transmission~\citep{mukhopadhyay2026coopertrim}. Recent methods further improve the perception--communication tradeoff through adaptive feature selection~\citep{xu2025cosdh, hu2026pragmatic, liu2026rate}. In the semantic communication literature, existing task-oriented methods tailor transmitted representations according to downstream task utility~\citep{xie2021deeplearningenabledsemantic, gunduz2022beyond}, jointly allocate communication resources under bandwidth constraints~\citep{zhang2023drl, liu2023adaptable, cai2025query}, and account for communication latency~\citep{sagduyu2026semanticcommunicationmeetsqueueing}.
However, these methods are primarily designed for individual perception or communication tasks, rather than the broader CPS setting where multiple downstream tasks with heterogeneous deadlines coexist and share a bandwidth-constrained communication link.


\section{Problem Formulation}
\label{sec:problem}

\paragraph{System Model.}
Following common settings in prior works~\citep{liu2020when2com, huwhere2comm, yang2023how2comm}, we consider cyber-physical agents that locally encode their observations and exchange latent representations over a communication link. For an observation $x_i$, a local encoder produces a latent representation:
\begin{equation}
H_i = E(x_i) = \{h_{i,1},\ldots,h_{i,N_i}\},
\end{equation}
where each $h_{i,j}\in\mathbb{R}^{d}$ is a fixed-dimensional latent token, and $N_i$ is the token length. Encoders may be modality-specific, e.g., for image, audio, or text inputs. Within each agent, multiple downstream tasks may coexist and generate latent messages that share the same communication link.

\paragraph{Network and Timing Model.}
We denote the available rate of the communication link by $R(t)$, which may vary over time. Within each agent, multiple latent messages from different tasks compete for the same communication resources. We denote the set of messages pending transmission at time $t$ by $\mathcal{Q}(t)$ and the deadline of message $i$ by $d_i$. Let $C_i$ denote the transmission completion time of message $i$. Following a common firm-deadline model in real-time systems, we assume that a message contributes to its downstream task only if the selected latent information arrives before its deadline, i.e., $C_i \leq d_i$.

\paragraph{Scheduling Objective.}
At time $t$, the scheduler considers all messages in the pending set $\mathcal{Q}(t)$. For each message $i\in\mathcal{Q}(t)$, let $K_i\leq N_i$ denote the number of latent tokens selected for transmission, where $K_i=0$ indicates that the message is not selected. We denote the semantic gain and communication cost of transmitting $K_i$ tokens by $U_i(K_i)$ and $B_i(K_i)$, respectively. Our goal is to determine which pending messages to transmit and how many tokens to transmit for each message, so as to maximize the total semantic gain delivered before their respective deadlines:
\begin{equation}
    \max_{\{K_i\}}
    \sum_{i\in\mathcal{Q}(t)}
    U_i(K_i)\mathbf{1}\!\left[C_i\leq d_i\right],
\end{equation}
subject to available bandwidth constraints.

\section{Approach}
\label{sec:approach}

\subsection{Overview}

\begin{figure*}[t]
    \centering
    \includegraphics[width=0.96\textwidth]{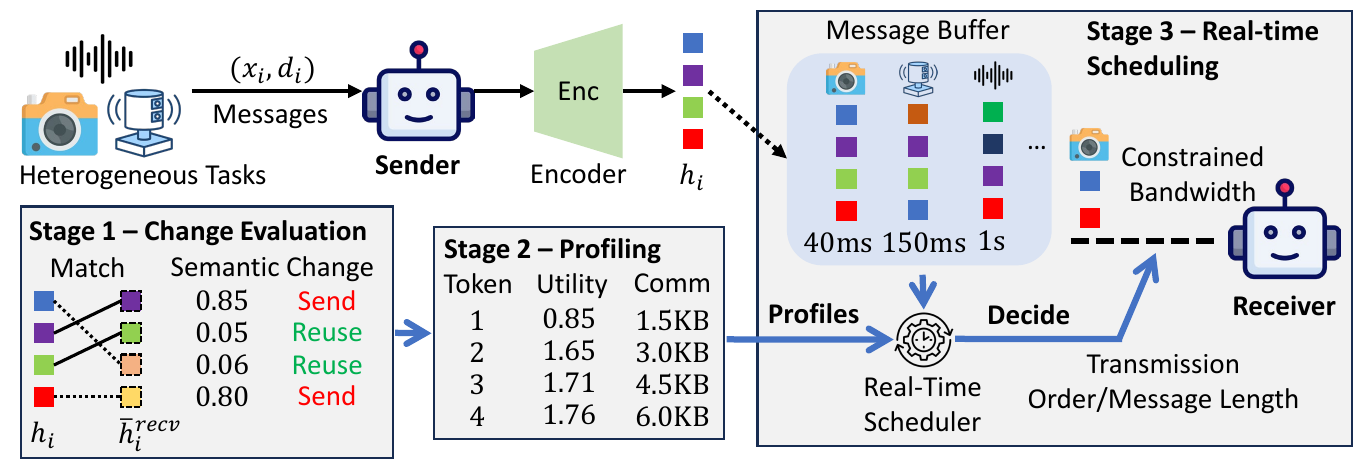}
    \caption{\sysname system design. Semantic change evaluation and utility--cost profiling guide the real-time scheduler in jointly selecting messages and transmission sizes under bandwidth and deadline constraints.}
    \label{fig:system-design} 
\end{figure*}

\sysname consists of three stages: \emph{semantic change evaluation}, \emph{utility--cost profiling}, and \emph{real-time scheduling}. In the first two stages, it identifies latent information that is new relative to the receiver's acknowledged state, and characterizes the semantic benefit and communication cost of different transmission lengths. The scheduler then combines these profiles with available bandwidth and task deadlines to determine which message to transmit and how many latent tokens to send. We demonstrate the system flow of \sysname in Fig. \ref{fig:system-design}

\subsection{Semantic Change Evaluation}

Physical observations often evolve continuously, creating substantial redundancy between consecutive observations. In this stage, we aim to measure such semantic redundancy and identify the nonredundant information to transmit. A key challenge is that directly comparing two latent messages token by token can be unreliable because their corresponding observations may be misaligned due to object or viewpoint motion. \sysname therefore first establishes token-level correspondences between the current latent message and the receiver's acknowledged latent state before quantifying their semantic changes.

Therefore, for each current token $h_{i,j}$, we first find the most similar token in the receiver's acknowledged latent state:
\begin{equation}
    \mu_i(j)=\arg\max_k\cos\!\left(h_{i,j},\tilde h^{\mathrm{recv}}_{i,k}\right).
\end{equation}
Let $c_{i,j}$ denote the corresponding maximum cosine similarity. We then quantify the semantic change of token $h_{i,j}$ as:
\begin{equation}
    s_{i,j}=1-c_{i,j}.
\end{equation}
Consequently, highly similar tokens receive small change scores, while tokens with low similarity receive larger scores. We further use a similarity threshold $\tau$ to determine whether the matched receiver-side token can be reused: a match is reusable only when $c_{i,j}\geq\tau$. Matching is performed against the receiver's acknowledged state to ensure that only information confirmed to be available at the receiver is reused.

\subsection{Utility--Cost Profiling}
The semantic change scores obtained above capture the amount of new information carried by individual latent tokens. In this stage, we further characterize the trade-off between semantic gain and communication cost at different transmission sizes $K$, where $K \leq N_i$ denotes the number of latent tokens transmitted for message $i$.

Specifically, we prioritize tokens with larger semantic change scores by sorting them in descending order according to $s_{i,j}$. We define the cumulative semantic gain and communication cost as:
\begin{equation}
\begin{aligned}
    U_i(K)&=\sum_{k=1}^{K} s_{i,(k)}, \\
    B_i(K)&=K b_{\mathrm{tok}}+B_i^{\mathrm{meta}}(K),
\end{aligned}
\end{equation}
where $s_{i,(k)}$ denotes the $k$-th largest semantic change score, $b_{\mathrm{tok}}$ is the serialized size of one latent token, and $B_i^{\mathrm{meta}}(K)$ accounts for the metadata required to interpret the transmission, including token indices, correspondence mappings, and the reference-state version. The resulting pairs $(U_i(K),B_i(K))$ characterize the available transmission choices for message $i$. As a result, a larger $K$ generally preserves more new semantic information but incurs greater communication cost.

\subsection{Real-Time Scheduling}

In the final stage, the scheduler takes as input the utility--cost profiles $\{(U_i(K),B_i(K))\}$ and task deadlines of all messages in $\mathcal{Q}(t)$, together with the current link rate $R(t)$. Based on this information, it determines the transmission size $K_i$ for each message and their transmission order.

\sysname first determines the transmission size $K_i$ for each pending message. Following the scheduling objective in Section~\ref{sec:problem}, we employ a greedy algorithm to maximize semantic gain under the deadline constraints. Specifically, we initialize $K_i=0$ for each pending message and iteratively allocate tokens one at a time. Suppose message $i$ has already been allocated $K_i$ tokens. The marginal semantic gain per communication cost of adding its next token $k=K_i+1$ is:
\begin{equation}
    \rho_{i,k}
    =\frac{U_i(k)-U_i(k-1)}
    {B_i(k)-B_i(k-1)}.
\end{equation}
At each step, the scheduler considers the next token of every pending message and prioritizes the one with the largest $\rho_{i,k}$, subject to the message deadlines. To check deadline feasibility, let $\pi_1,\ldots,\pi_m$ denote the currently selected messages ordered by earliest deadline first (EDF). The completion time of message $\pi_q$ is estimated as:
\begin{equation}
    C_{\pi_q}
    =t+\sum_{\ell=1}^{q}
    \frac{8B_{\pi_\ell}(K_{\pi_\ell})}{R(t)}.
\end{equation}
An allocation is feasible only if $C_{\pi_q}\leq d_{\pi_q}$ for all $q=1,\ldots,m$. Among all feasible token increments, the scheduler selects the one with the largest $\rho_{i,k}$, updates $K_i\leftarrow K_i+1$, and repeats this process until no additional token can be allocated without violating a deadline.

The resulting $K_i$ values determine the transmission size of each message, where $K_i=0$ indicates that the message is not selected. Messages with $K_i>0$ are transmitted in EDF order. In this way, \sysname prioritizes semantic gain relative to communication cost while using task deadlines as feasibility constraints on both message selection and transmission size. After transmission, the receiver updates its latent state with the received tokens and reuses matched tokens from its previous state according to the correspondences established in the semantic change evaluation stage. The receiver then acknowledges the updated state, allowing the sender to maintain an accurate view of the receiver's state for subsequent communication. The detailed scheduling algorithm can be found in Appendix \ref{alg:scheduling}.

\section{Evaluation}
\label{sec:evaluation}

\subsection{Experimental Settings}
\label{sec:eval-setup}

\paragraph{CPS Benchmarks.}
We evaluate \sysname on three cyber-physical benchmarks: CooperScene~\cite{wu2026cooperscene}, OPV2V~\cite{xu2022opv2v}, and LIBERO~\cite{liu2023libero}. CooperScene is a real-world cooperative autonomous-driving benchmark with LiDAR and camera observations from vehicles and roadside infrastructure. OPV2V is a simulated cooperative-perception benchmark with LiDAR and camera observations from multiple vehicles. LIBERO provides language-conditioned robotic manipulation tasks with visual observations and robot states. These benchmarks cover different CPS use cases, and their detailed descriptions are provided in Appendix \ref{sec:settings}.

\paragraph{Tasks and Models.} On CooperScene, we evaluate concurrent 3D object detection and 5-sec trajectory prediction tasks, with deadlines of 100 ms and 200 ms. We use CoBEVT \cite{xu2022cobevt} for detection and a frozen, pretrained Motion Transformer (MTR) \cite{shi2022motion} for prediction. On OPV2V, we evaluate concurrent 3D object detection and BEV semantic segmentation using the LiDAR-based detection and camera-based segmentation variants of CoBEVT, each with its own pretrained weights. Their deadlines are 100 ms and 200 ms, respectively. On LIBERO, we evaluate object manipulation using SmolVLA \cite{shukor2025smolvla}, with the deadline determined by the time remaining before the robot exhausts its buffered actions, plus reserving 100 ms for inference and 10 ms for returning actions. \textit{Across all benchmarks, a sensing agent sends latent features to another agent or remote server for inference.} 

\paragraph{Network Settings.} For each benchmark, we evaluate four fixed-bandwidth settings and one time-varying setting. The fixed rates correspond to different fractions (e.g., 20\%, 50\%) of the average bandwidth required for full latent transmission in each benchmark. In the time-varying setting, the link alternates between low and high rates every 2 seconds. In all settings, we account for both queueing and transmission time when computing communication delay. More detailed network settings, including the bandwidth values for each benchmark, are provided in Table~\ref{tab:benchmark-settings} in the Appendix.

\begin{table*}[t]
    \centering
    \footnotesize
    \setlength{\tabcolsep}{2pt}
    \renewcommand{\arraystretch}{1.05}
    \begin{tabular*}{\textwidth}
        {@{\extracolsep{\fill}}l*{10}{r}@{}}
        \toprule
        \multicolumn{11}{l}{\textbf{(a) Detection}} \\
        Method
        & \multicolumn{2}{c}{15 Mbps}
        & \multicolumn{2}{c}{20 Mbps}
        & \multicolumn{2}{c}{25 Mbps}
        & \multicolumn{2}{c}{30 Mbps}
        & \multicolumn{2}{c}{Varying} \\
        \cmidrule(lr){2-3}
        \cmidrule(lr){4-5}
        \cmidrule(lr){6-7}
        \cmidrule(lr){8-9}
        \cmidrule(lr){10-11}
        & AP $\uparrow$ & OT $\uparrow$ & AP $\uparrow$ & OT $\uparrow$ & AP $\uparrow$ & OT $\uparrow$ & AP $\uparrow$ & OT $\uparrow$ & AP $\uparrow$ & OT $\uparrow$ \\
        \midrule
        Full-send
        & 22.77 & 0.00
        & 22.77 & 0.00
        & 22.77 & 0.00
        & 22.77 & 0.00
        & 40.26 & 49.97 \\
        Uniform
        & 22.50 & \textbf{68.42}
        & 23.38 & 68.64
        & 23.72 & 68.63
        & 24.91 & 68.50
        & 39.25 & 81.59 \\
        CoSDH
        & 27.99 & 32.04
        & 31.34 & 48.02
        & 40.67 & 68.14
        & 43.33 & 75.43
        & 49.87 & 89.05 \\
        V2X-ViT
        & 25.24 & 0.00
        & 25.24 & 0.00
        & 25.33 & 1.21
        & 25.91 & 8.70
        & 42.22 & 55.01 \\
        \textsc{Scale}
        & \textbf{45.62} & 64.95
        & \textbf{51.53} & \textbf{80.36}
        & \textbf{54.87} & \textbf{88.13}
        & \textbf{56.28} & \textbf{92.84}
        & \textbf{56.09} & \textbf{94.52} \\
        \bottomrule
    \end{tabular*}

    \par\vspace{4pt}

    \begin{tabular*}{\textwidth}
        {@{\extracolsep{\fill}}l*{10}{r}@{}}
        \toprule
        \multicolumn{11}{l}{\textbf{(b) Prediction}} \\
        Method
        & \multicolumn{2}{c}{15 Mbps}
        & \multicolumn{2}{c}{20 Mbps}
        & \multicolumn{2}{c}{25 Mbps}
        & \multicolumn{2}{c}{30 Mbps}
        & \multicolumn{2}{c}{Varying} \\
        \cmidrule(lr){2-3}
        \cmidrule(lr){4-5}
        \cmidrule(lr){6-7}
        \cmidrule(lr){8-9}
        \cmidrule(lr){10-11}
        & Succ. $\uparrow$ & OT $\uparrow$
        & Succ. $\uparrow$ & OT $\uparrow$
        & Succ. $\uparrow$ & OT $\uparrow$
        & Succ. $\uparrow$ & OT $\uparrow$
        & Succ. $\uparrow$ & OT $\uparrow$ \\
        \midrule
        Full-send
        & 43.72 & 0.26
        & 44.73 & 6.54
        & 49.57 & 35.32
        & 48.50 & 25.05
        & 54.88 & 64.89 \\
        Uniform
        & 59.49 & 93.81
        & 59.52 & 94.44
        & 59.87 & 95.64
        & 59.84 & 95.90
        & 60.48 & 98.47 \\
        CoSDH
        & 46.48 & 16.69
        & 51.28 & 44.26
        & 52.86 & 50.99
        & 57.37 & 80.92
        & 59.24 & 91.47 \\
        V2X-ViT
        & 48.54 & 29.23
        & 46.65 & 18.53
        & 49.75 & 36.37
        & 52.41 & 50.95
        & 57.07 & 77.64 \\
        \textsc{Scale}
        & \textbf{60.82} & \textbf{99.82}
        & \textbf{60.85} & \textbf{100.00}
        & \textbf{60.85} & \textbf{100.00}
        & \textbf{60.85} & \textbf{100.00}
        & \textbf{60.81} & \textbf{99.71} \\
        \bottomrule
    \end{tabular*}
    \caption{CooperScene results on 3,936 detection jobs and 18,693 prediction targets. AP denotes average precision, OT denotes the on-time rate (\%), and Succ. denotes prediction success (\%).}
    \label{tab:cooperscene}
\end{table*}

\begin{table*}[t]
    \centering
    \footnotesize
    \setlength{\tabcolsep}{2pt}
    \renewcommand{\arraystretch}{1.05}
    \begin{tabular*}{\textwidth}
        {@{\extracolsep{\fill}}l*{10}{r}@{}}
        \toprule
        \multicolumn{11}{l}{\textbf{(a) Detection}} \\
        Method
        & \multicolumn{2}{c}{8 Mbps}
        & \multicolumn{2}{c}{15 Mbps}
        & \multicolumn{2}{c}{31 Mbps}
        & \multicolumn{2}{c}{61 Mbps}
        & \multicolumn{2}{c}{Varying} \\
        \cmidrule(lr){2-3}
        \cmidrule(lr){4-5}
        \cmidrule(lr){6-7}
        \cmidrule(lr){8-9}
        \cmidrule(lr){10-11}
        & AP $\uparrow$ & OT $\uparrow$ & AP $\uparrow$ & OT $\uparrow$ & AP $\uparrow$ & OT $\uparrow$ & AP $\uparrow$ & OT $\uparrow$ & AP $\uparrow$ & OT $\uparrow$ \\
        \midrule
        Full-send
        & 76.37 & 0.00
        & 76.37 & 0.00
        & 76.37 & 0.00
        & 76.37 & 0.00
        & 76.37 & 0.00 \\
        Uniform
        & 78.29 & \textbf{46.49}
        & 79.87 & 44.21
        & 77.80 & 41.20
        & 81.90 & 39.93
        & 79.65 & 42.30 \\
        CooperTrim
        & \textbf{84.25} & 0.73
        & 84.39 & 4.92
        & 85.81 & 28.71
        & 89.47 & 69.55
        & 86.14 & 31.81 \\
        Where2comm
        & 82.25 & 0.82
        & 82.63 & 10.30
        & 88.09 & 54.42
        & 91.88 & 86.51
        & 86.20 & 36.37 \\
        \textsc{Scale}
        & 83.41 & 37.74
        & \textbf{86.42} & \textbf{50.23}
        & \textbf{90.30} & \textbf{75.39}
        & \textbf{92.52} & \textbf{91.70}
        & \textbf{86.44} & \textbf{56.43} \\
        \bottomrule
    \end{tabular*}

    \par\vspace{4pt}

    \begin{tabular*}{\textwidth}
        {@{\extracolsep{\fill}}l*{10}{r}@{}}
        \toprule
        \multicolumn{11}{l}{\textbf{(b) Segmentation}} \\
        Method
        & \multicolumn{2}{c}{8 Mbps}
        & \multicolumn{2}{c}{15 Mbps}
        & \multicolumn{2}{c}{31 Mbps}
        & \multicolumn{2}{c}{61 Mbps}
        & \multicolumn{2}{c}{Varying} \\
        \cmidrule(lr){2-3}
        \cmidrule(lr){4-5}
        \cmidrule(lr){6-7}
        \cmidrule(lr){8-9}
        \cmidrule(lr){10-11}
        & mIoU $\uparrow$ & OT $\uparrow$ & mIoU $\uparrow$ & OT $\uparrow$ & mIoU $\uparrow$ & OT $\uparrow$
        & mIoU $\uparrow$ & OT $\uparrow$ & mIoU $\uparrow$ & OT $\uparrow$ \\
        \midrule
        Full-send
        & 27.02 & 0.00
        & 27.14 & 0.73
        & 28.80 & 10.36
        & 31.77 & 27.82
        & 28.74 & 11.09 \\
        Uniform
        & 27.10 & 77.82
        & 26.95 & 81.09
        & 28.93 & 79.27
        & 32.69 & 78.00
        & 28.67 & 76.18 \\
        CooperTrim
        & 33.08 & 12.36
        & 35.52 & 32.36
        & 40.87 & 67.64
        & 43.31 & 92.73
        & 37.75 & 48.55 \\
        Where2comm
        & 37.14 & 0.18
        & 37.17 & 7.82
        & 37.76 & 52.00
        & 38.32 & 93.64
        & 37.56 & 42.55 \\
        \textsc{Scale}
        & \textbf{46.48} & \textbf{94.91}
        & \textbf{47.14} & \textbf{97.09}
        & \textbf{47.71} & \textbf{99.64}
        & \textbf{47.78} & \textbf{100.00}
        & \textbf{45.17} & \textbf{90.18} \\
        \bottomrule
    \end{tabular*}
    \caption{OPV2V results on 528 detection jobs and
    265 segmentation jobs. AP denotes average precision, mIoU denotes mean IoU score, and OT denotes on-time rate (\%).}
    \label{tab:opv2v}
\end{table*}

\paragraph{Baselines.}
We compare \sysname with \emph{Full-send}, which transmits the complete latent message, and \emph{Uniform compression}, which uniformly selects a fraction of latent features for transmission based on the available bandwidth. On CooperScene, we additionally compare with CoSDH~\citep{xu2025cosdh} and V2X-ViT~\citep{xu2022v2x}. On OPV2V, we additionally compare with CooperTrim~\citep{mukhopadhyay2026coopertrim} and Where2comm~\citep{huwhere2comm}. All baseline methods use earliest-deadline-first (EDF) transmission ordering. 

\paragraph{Metrics.}
We evaluate \sysname in terms of task utility and message on-time rate. Task utility measures downstream task performance, while on-time rate is the percentage of messages received before their deadlines.

\subsection{Benchmark Performance}
\label{sec:benchmark-performance}

\paragraph{CooperScene.}
We use task-specific performance metrics to evaluate the two tasks on CooperScene. For the 3D object detection task, we report average precision (AP) and the message on-time rate, where AP is computed using an intersection-over-union (IoU) threshold of 0.5 to match predicted boxes to ground-truth boxes. For the trajectory prediction task, we report task success and the message on-time rate, where task success is the percentage of prediction targets for which at least one of six predicted trajectories is within 2~m of the ground-truth position.

Table~\ref{tab:cooperscene} presents the results. The results show that \sysname achieves the highest detection AP and prediction task success across all settings. In particular, for 3D object detection at 20~Mbps, \sysname achieves 51.53 AP, exceeding the strongest baseline, CoSDH, by 20.19 points. It also delivers 80.36\% of detection messages on time, compared with 48.02\% for CoSDH. For trajectory prediction under fixed bandwidth, \sysname maintains task success between 60.82\% and 60.85\%, while delivering at least 99.82\% of prediction messages on time. Under varying bandwidth, \sysname achieves 56.09 detection AP and 60.81\% prediction task success, with detection and prediction on-time rates of 94.52\% and 99.71\%, respectively.

\paragraph{OPV2V.}
Similarly, we use task-specific performance metrics to evaluate the two tasks on OPV2V. For the 3D object detection task, we report AP and the message on-time rate. We evaluate BEV semantic segmentation using mean intersection-over-union (mIoU) and also report its message on-time rate, where mIoU averages the IoU between predicted and ground-truth regions across semantic classes.

Table~\ref{tab:opv2v} presents the results. The results show that \sysname achieves the highest segmentation mIoU and segmentation on-time rate across all settings, and the highest detection AP except at 8~Mbps. In this setting, \sysname achieves 83.41 AP, slightly below CooperTrim's 84.25 AP, while delivering 37.74\% of detection messages on time, compared with 0.73\% for CooperTrim. For BEV semantic segmentation, the largest mIoU improvement over the strongest baseline occurs at 15~Mbps, where \sysname achieves 47.14 mIoU, exceeding Where2comm by 9.97 points. It also delivers 97.09\% of segmentation messages on time, compared with 7.82\% for Where2comm. 

\begin{table}[!tp]
    \centering
    \small
    \setlength{\tabcolsep}{4pt}
    \begin{tabular}{@{}llrr@{}}
    \hline
    Bandwidth & Method & Task success $\uparrow$ & On-time $\uparrow$ \\
    \hline
    \multirow{3}{*}{3.4 Mbps} & Full-send & 0.00\% & 0.00\% \\
     & Uniform & 0.00\% & 74.47\% \\
     & \sysname & \textbf{44.44\%} & \textbf{77.76\%} \\
    \hline
    \multirow{3}{*}{5.0 Mbps} & Full-send & 24.44\% & 25.24\% \\
     & Uniform & 0.00\% & 74.47\% \\
     & \sysname & \textbf{86.67\%} & \textbf{84.57\%} \\
    \hline
    \multirow{3}{*}{6.7 Mbps} & Full-send & 60.00\% & 25.98\% \\
     & Uniform & 0.00\% & 74.47\% \\
     & \sysname & \textbf{95.56\%} & \textbf{88.81\%} \\
    \hline
    \multirow{3}{*}{10.1 Mbps} & Full-send & 88.89\% & 51.52\% \\
     & Uniform & 42.22\% & 74.85\% \\
     & \sysname & \textbf{97.78\%} & \textbf{94.21\%} \\
    \hline
    \multirow{3}{*}{Varying} & Full-send & 88.89\% & 65.87\% \\
     & Uniform & 0.00\% & 77.45\% \\
     & \sysname & \textbf{100.00\%} & \textbf{89.28\%} \\
    \hline
    \end{tabular}
    \caption{LIBERO results over 45 episodes per method and bandwidth setting. We report the task success rate (\%) and on-time rate (\%).}
    \label{tab:scale-libero}
\end{table}

\begin{figure}[t]
    \centering
    \includegraphics[width=0.49\textwidth]{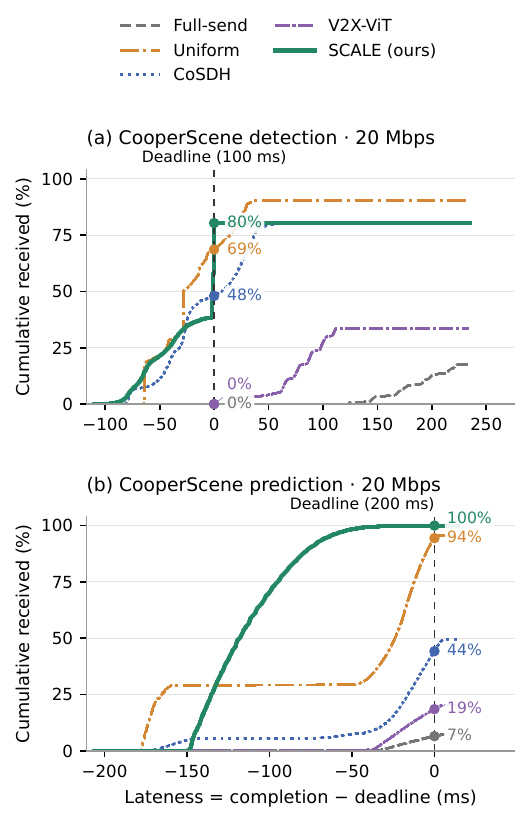}
    \caption{Message-lateness CDFs for CooperScene at 20~Mbps: (a) object detection and (b) trajectory prediction. The vertical dashed line marks each task's deadline; the cumulative percentage at zero gives the on-time rate.} 
    \label{fig:cdf-cooperscene-20mbps}
\end{figure}

\paragraph{LIBERO.}
Our LIBERO evaluation focuses on a single language-conditioned robotic manipulation task. We report task success, defined as the percentage of episodes completed within 280 steps, and the communication on-time rate.

Table~\ref{tab:scale-libero} presents the results. The results show that \sysname achieves the highest task success and on-time rate across all settings. At the lowest tested bandwidth of 3.4~Mbps, \sysname completes 44.44\% of episodes, while both baselines fail all episodes. In particular, Uniform compression achieves a similar on-time rate (74.47\% versus 77.76\% for \sysname), showing that timely delivery alone does not ensure task success. The largest task success improvement over the strongest baseline occurs at 5.0~Mbps, where \sysname achieves 86.67\% task success, compared with 24.44\% for Full-send and 0.00\% for Uniform compression. Its communication on-time rate also reaches 84.57\%, exceeding both baselines.

\subsection{Real-Time Performance}
\label{sec:realtime-performance}

We measure message lateness as $L_i=C_i-d_i$, where $C_i$ and $d_i$ are the completion time and deadline of message $i$, respectively. On-time messages have $L_i\leq 0$, while delayed messages have $L_i>0$. To better illustrate real-time performance, Fig.~\ref{fig:cdf-cooperscene-20mbps} shows the lateness distributions for CooperScene at 20~Mbps for both object detection and trajectory prediction tasks. In the figure, the cumulative percentage at zero gives the on-time rate, which is computed over all generated messages, including undelivered ones.

The results show that \sysname delivers 80.36\% of detection messages and 100.00\% of prediction messages on time, compared with 68.64\% and 94.44\% for Uniform compression. Uniform compression achieves the highest on-time rates among the baselines at this bandwidth, but its task utility is significantly lower than that of \sysname. We also observe a steep rise in the detection CDF just before the deadline for the object detection task, indicating that many messages complete shortly before their deadlines. This behavior is consistent with \sysname's greedy allocation of available bandwidth to maximize semantic gain under deadline constraints. The complete distributions for all three benchmarks and bandwidth settings appear in Appendix~\ref{app:additional-results} (Figures~\ref{fig:cdf-cooperscene-selected}, \ref{fig:cdf-opv2v-selected}, and~\ref{fig:cdf-libero-selected}). Across these results, \sysname improves on-time delivery in every tested setting and achieves the highest on-time rate among all methods in most settings.

\subsection{Case Study}
\label{sec:case-study}

\begin{figure}[t]
    \centering
    \includegraphics[width=\linewidth]{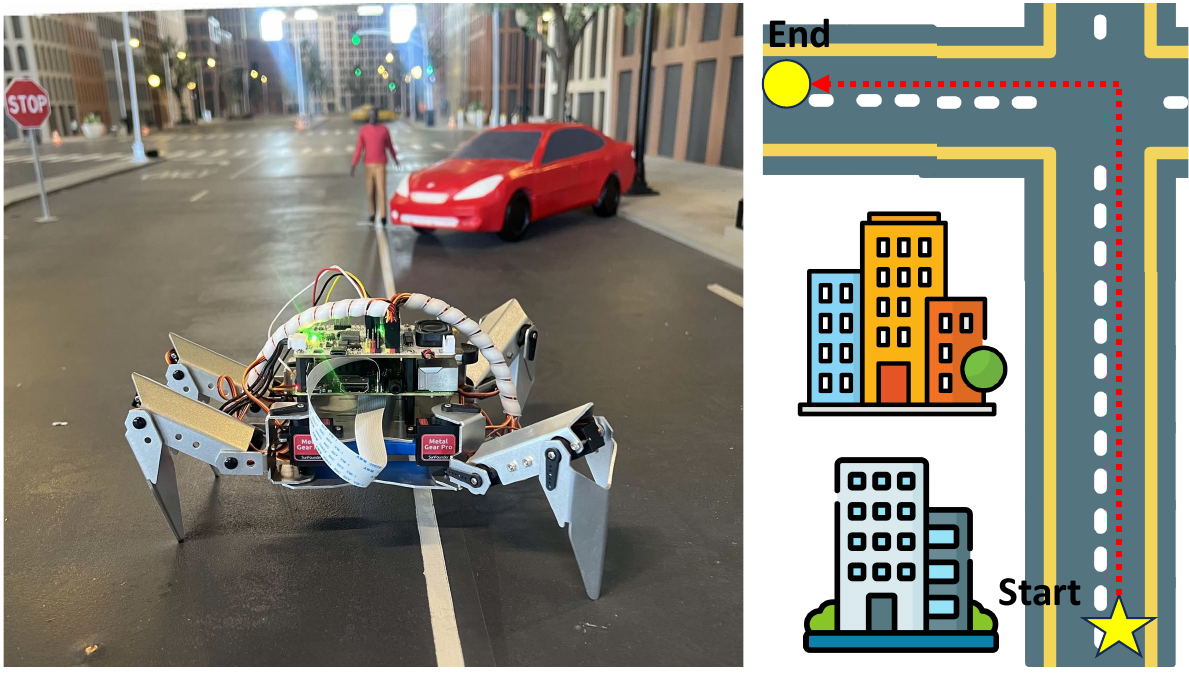}
    \caption{Spider robot patrolling in the mini-city with pedestrians, stop signs, and model cars. The spider patrols along an L-shaped route.}
    \label{fig:spider-case}
\end{figure}

We conduct a hardware-in-the-loop case study using a Raspberry Pi-based spider robot, as shown in Fig.~\ref{fig:spider-case}. The robot is equipped with a camera and an ultrasonic sensor and performs two tasks: object detection and distance measurement on a fixed robot patrolling route. The patrolling route is a 55.4-second walk through a miniature city containing objects such as small road signs, pedestrian figures, and buildings. Among the two tasks, object detection is time-sensitive, with a communication deadline of 1,000~ms, while ultrasonic distance measurements generate background traffic over the shared link. We compare \sysname against two baselines: Raw RGB, which transmits uncompressed images, and Full-send, which transmits complete intermediate latent features. We evaluate the system in three bandwidth settings, including 1, 2, and 8~Mbps.


\begin{table}[t]
    \centering
    \small
    \setlength{\tabcolsep}{4pt}
    \begin{tabular}{@{}llrrr@{}}
        \toprule
        & & \multicolumn{3}{c}{Bandwidth (Mbps)} \\
        \cmidrule(l){3-5}
        Method & Metric & 1 & 2 & 8 \\
        \midrule
        \multirow{2}{*}{RGB}
        & Success $\uparrow$ & 0.00\% & 0.00\% & 100.00\% \\
        & On-time $\uparrow$ & 0.00\% & 0.00\% & 100.00\% \\
        \midrule
        \multirow{2}{*}{Latent}
        & Success $\uparrow$ & 0.00\% & 0.00\% & 100.00\% \\
        & On-time $\uparrow$ & 0.00\% & 0.00\% & 100.00\% \\
        \midrule
        \multirow{2}{*}{SCALE}
        & Success $\uparrow$ & \textbf{77.62\%} & \textbf{91.67\%} & 100.00\% \\
        & On-time $\uparrow$ & \textbf{100.00\%} & \textbf{96.43\%} & 100.00\% \\
        \bottomrule
    \end{tabular}
    \caption{Spider robot mini-city case study results under different bandwidth constraints.}
    \label{tab:spider-case}
\end{table}

We report task success and message on-time rates. Table~\ref{tab:spider-case} presents the results. We observe that \sysname substantially outperforms both baselines under constrained bandwidth. Specifically, at 1 and 2~Mbps, both baselines achieve zero detection success, whereas \sysname achieves success rates of 77.62\% and 91.67\%, respectively, with on-time rates of 100\% and 96.43\%. At 8~Mbps, all methods achieve 100\% detection success and on-time delivery. These results demonstrate the effectiveness of \sysname in preserving timely detection performance on a physical robotic platform under limited bandwidth.


\section{Conclusion}

In this work, we present \sysname, a real-time latent communication method for cyber-physical agents. By exploiting spatiotemporal continuity, \sysname avoids retransmitting latent information already available at the receiver. To support timely communication, we design a real-time scheduler that jointly selects messages and transmission sizes based on semantic gain, available bandwidth, and task deadlines. We evaluate \sysname on three benchmarks and conduct a physical robot case study. The results demonstrate improved task performance and high message on-time rates under constrained and varying network conditions. These findings highlight \sysname as an effective and efficient method for meeting the communication and timing requirements of CPS agents.

\section*{Limitations}
While \sysname significantly improves bandwidth efficiency and supports real-time scheduling for CPS agents, it has two limitations. First, the communication savings of \sysname depend on spatiotemporal redundancy in latent representations. However, observations in the physical world can change abruptly, which may reduce the amount of reusable information and increase the communication required to maintain task performance. Second, the scheduling algorithm assesses deadline feasibility using the current link rate. However, this design cannot handle unexpected bandwidth decreases that occur on timescales shorter than the minimum scheduling interval, which may affect message timeliness before the scheduler can adapt.

\FloatBarrier
\bibliography{reference}

@inproceedings{xu2025cosdh,
  title={Cosdh: Communication-efficient collaborative perception via supply-demand awareness and intermediate-late hybridization},
  author={Xu, Junhao and Zhang, Yanan and Cai, Zhi and Huang, Di},
  booktitle={2025 IEEE/CVF Conference on Computer Vision and Pattern Recognition (CVPR)},
  pages={6834--6843},
  year={2025},
  organization={IEEE}
}

@inproceedings{xu2022v2x,
  title={V2x-vit: Vehicle-to-everything cooperative perception with vision transformer},
  author={Xu, Runsheng and Xiang, Hao and Tu, Zhengzhong and Xia, Xin and Yang, Ming-Hsuan and Ma, Jiaqi},
  booktitle={European conference on computer vision},
  pages={107--124},
  year={2022},
  organization={Springer}
}

@inproceedings{du2026interlat,
  title     = {Enabling Agents to Communicate Entirely in Latent Space},
  author    = {Du, Zhuoyun and Wang, Runze and Bai, Huiyu and Cao, Zouying and Zhu, Xiaoyong and Cheng, Yu and Zheng, Bo and Chen, Wei and Ying, Haochao},
  booktitle = {Proceedings of the 64th Annual Meeting of the Association for Computational Linguistics (Volume 1: Long Papers)},
  pages     = {27106--27129},
  address   = {San Diego, California, United States},
  publisher = {Association for Computational Linguistics},
  month     = jul,
  year      = {2026},
  doi       = {10.18653/v1/2026.acl-long.1248},
  url       = {https://aclanthology.org/2026.acl-long.1248/}
}

@inproceedings{shi2026selectivekvcomm,
  title     = {{KVComm}: Enabling Efficient {LLM} Communication through Selective {KV} Sharing},
  author    = {Shi, Xiangyu and Chiesa, Marco and Maguire, Jr., Gerald Q. and Kosti{\'c}, Dejan},
  booktitle = {The Fourteenth International Conference on Learning Representations},
  year      = {2026},
  url       = {https://proceedings.iclr.cc/paper_files/paper/2026/hash/883105b282fe15275991b411e6b200c5-Abstract-Conference.html}
}

@inproceedings{ye2025crosscontextkvcomm,
  title     = {{KVCOMM}: Online Cross-Context {KV}-Cache Communication for Efficient {LLM}-Based Multi-Agent Systems},
  author    = {Ye, Hancheng and Gao, Zhengqi and Ma, Mingyuan and Wang, Qinsi and Fu, Yuzhe and Chung, Ming-Yu and Lin, Yueqian and Liu, Zhijian and Zhang, Jianyi and Zhuo, Danyang and Chen, Yiran},
  booktitle = {Advances in Neural Information Processing Systems},
  volume    = {38},
  publisher = {Curran Associates, Inc.},
  year      = {2025},
  doi       = {10.52202/085713-0605},
  url       = {https://proceedings.neurips.cc/paper_files/paper/2025/hash/1a074a28c3a6f2056562d00649ae6416-Abstract-Conference.html}
}

@inproceedings{zou2026latentmas,
  title     = {Latent Collaboration in Multi-Agent Systems},
  author    = {Zou, Jiaru and Yang, Xiyuan and Qiu, Ruizhong and Li, Gaotang and Tieu, Katherine and Lu, Pan and Shen, Ke and Tong, Hanghang and Choi, Yejin and He, Jingrui and Zou, James and Wang, Mengdi and Yang, Ling},
  booktitle = {Proceedings of the 43rd International Conference on Machine Learning},
  year      = {2026}
}

@misc{sagduyu2026semanticcommunicationmeetsqueueing,
  title         = {When Semantic Communication Meets Queueing: Cross-Layer Latency and Task Fidelity Optimization},
  author        = {Sagduyu, Yalin E. and Erpek, Tugba},
  year          = {2026},
  eprint        = {2605.05514},
  archiveprefix = {arXiv},
  primaryclass  = {cs.IT},
  url           = {https://arxiv.org/abs/2605.05514}
}

@misc{xie2021deeplearningenabledsemantic,
  title         = {Deep Learning Enabled Semantic Communication Systems},
  author        = {Xie, Huiqiang and Qin, Zhijin and Li, Geoffrey Ye and Juang, Biing-Hwang},
  year          = {2021},
  eprint        = {2006.10685},
  archiveprefix = {arXiv},
  primaryclass  = {eess.SP},
  doi           = {10.1109/TSP.2021.3071210},
  url           = {https://arxiv.org/abs/2006.10685}
}

@InProceedings{Patel_2021_ICCV,
    author    = {Patel, Shivansh and Wani, Saim and Jain, Unnat and Schwing, Alexander G. and Lazebnik, Svetlana and Savva, Manolis and Chang, Angel X.},
    title     = {Interpretation of Emergent Communication in Heterogeneous Collaborative Embodied Agents},
    booktitle = {Proceedings of the IEEE/CVF International Conference on Computer Vision (ICCV)},
    month     = {October},
    year      = {2021},
    pages     = {15953-15963}
}

@inproceedings{huwhere2comm,
 author = {Hu, Yue and Fang, Shaoheng and Lei, Zixing and Zhong, Yiqi and Chen, Siheng},
 booktitle = {Advances in Neural Information Processing Systems},
 doi = {10.52202/068431-0352},
 editor = {S. Koyejo and S. Mohamed and A. Agarwal and D. Belgrave and K. Cho and A. Oh},
 pages = {4874--4886},
 publisher = {Curran Associates, Inc.},
 title = {Where2comm: Communication-Efficient Collaborative Perception via Spatial Confidence Maps},
 url = {https://proceedings.neurips.cc/paper_files/paper/2022/file/1f5c5cd01b864d53cc5fa0a3472e152e-Paper-Conference.pdf},
 volume = {35},
 year = {2022}
}

@inproceedings{liu2020when2com,
  title={When2com: Multi-agent perception via communication graph grouping},
  author={Liu, Yen-Cheng and Tian, Junjiao and Glaser, Nathaniel and Kira, Zsolt},
  booktitle={2020 IEEE/CVF Conference on Computer Vision and Pattern Recognition (CVPR)},
  pages={4105--4114},
  year={2020},
  organization={IEEE}
}

@article{yang2023how2comm,
  title={How2comm: Communication-efficient and collaboration-pragmatic multi-agent perception},
  author={Yang, Dingkang and Yang, Kun and Wang, Yuzheng and Liu, Jing and Xu, Zhi and Yin, Rongbin and Zhai, Peng and Zhang, Lihua},
  journal={Advances in Neural Information Processing Systems},
  volume={36},
  pages={25151--25164},
  year={2023}
}

@inproceedings{hu2024communication,
  title={Communication-efficient collaborative perception via information filling with codebook},
  author={Hu, Yue and Peng, Juntong and Liu, Sifei and Ge, Junhao and Liu, Si and Chen, Siheng},
  booktitle={2024 IEEE/CVF Conference on Computer Vision and Pattern Recognition (CVPR)},
  pages={15481--15490},
  year={2024},
  organization={IEEE}
}

@inproceedings{
  mukhopadhyay2026coopertrim,
  title={{COOPERTRIM}: Adaptive Data Selection for Uncertainty-Aware Cooperative Perception},
  author={Shilpa Mukhopadhyay and Amit Roy-Chowdhury and Hang Qiu},
  booktitle={The Fourteenth International Conference on Learning Representations},
  year={2026}
}

@inproceedings{fu2026cache,
  title={Cache-to-cache: Direct semantic communication between large language models},
  author={Fu, Tianyu and Min, Zihan and Zhang, Hanling and Yan, Jichao and Dai, Guohao and Ouyang, Wanli and Wang, Yu},
  booktitle={International Conference on Learning Representations},
  volume={2026},
  pages={43130--43158},
  year={2026}
}

@inproceedings{chen2026condenseflow,
  title={CondenseFlow: Scalable Latent Space Collaboration via Semantic Compression for Multi-Agent Systems},
  author={Chen, Xiaoyu and Wu, Fengge and Zhao, Junsuo and Fan, Yun},
  booktitle={Findings of the Association for Computational Linguistics: ACL 2026},
  pages={13694--13712},
  year={2026}
}

@inproceedings{wang2020v2vnet,
  title={V2vnet: Vehicle-to-vehicle communication for joint perception and prediction},
  author={Wang, Tsun-Hsuan and Manivasagam, Sivabalan and Liang, Ming and Yang, Bin and Zeng, Wenyuan and Urtasun, Raquel},
  booktitle={European conference on computer vision},
  pages={605--621},
  year={2020},
  organization={Springer}
}

@article{li2021disconet,
  title={Learning distilled collaboration graph for multi-agent perception},
  author={Li, Yiming and Ren, Shunli and Wu, Pengxiang and Chen, Siheng and Feng, Chen and Zhang, Wenjun},
  journal={Advances in Neural Information Processing Systems},
  volume={34},
  pages={29541--29552},
  year={2021}
}

@inproceedings{lei2022latency,
  title={Latency-aware collaborative perception},
  author={Lei, Zixing and Ren, Shunli and Hu, Yue and Zhang, Wenjun and Chen, Siheng},
  booktitle={European Conference on Computer Vision},
  pages={316--332},
  year={2022},
  organization={Springer}
}

@article{wei2023asynchrony,
  title={Asynchrony-robust collaborative perception via bird's eye view flow},
  author={Wei, Sizhe and Wei, Yuxi and Hu, Yue and Lu, Yifan and Zhong, Yiqi and Chen, Siheng and Zhang, Ya},
  journal={Advances in Neural Information Processing Systems},
  volume={36},
  pages={28462--28477},
  year={2023}
}

@article{hu2026pragmatic,
  title={Pragmatic communication in multi-agent collaborative perception},
  author={Hu, Yue and Pang, Xianghe and Qin, Xiaoqi and Eldar, Yonina C and Chen, Siheng and Zhang, Ping and Zhang, Wenjun},
  journal={IEEE Transactions on Pattern Analysis and Machine Intelligence},
  year={2026},
  publisher={IEEE}
}

@inproceedings{liu2026rate,
  title={RATE-DISTORTION OPTIMIZED PRAGMATIC COMMUNICATION FOR COLLABORATIVE PERCEPTION},
  author={Liu, Genjia and Hu, Anning and Hu, Yue and Zhang, Wenjun and Chen, Siheng},
  booktitle={International Conference on Learning Representations},
  volume={2026},
  pages={119998--120019},
  year={2026}
}

@article{gunduz2022beyond,
  title={Beyond transmitting bits: Context, semantics, and task-oriented communications},
  author={G{\"u}nd{\"u}z, Deniz and Qin, Zhijin and Aguerri, Inaki Estella and Dhillon, Harpreet S and Yang, Zhaohui and Yener, Aylin and Wong, Kai Kit and Chae, Chan-Byoung},
  journal={IEEE Journal on Selected Areas in Communications},
  volume={41},
  number={1},
  pages={5--41},
  year={2022},
  publisher={IEEE}
}

@article{zhang2023drl,
  title={DRL-driven dynamic resource allocation for task-oriented semantic communication},
  author={Zhang, Haijun and Wang, Hongyu and Li, Yabo and Long, Keping and Nallanathan, Arumugam},
  journal={IEEE Transactions on Communications},
  volume={71},
  number={7},
  pages={3992--4004},
  year={2023},
  publisher={IEEE}
}

@article{liu2023adaptable,
  title={Adaptable semantic compression and resource allocation for task-oriented communications},
  author={Liu, Chuanhong and Guo, Caili and Yang, Yang and Jiang, Nan},
  journal={IEEE Transactions on Cognitive Communications and Networking},
  volume={10},
  number={3},
  pages={769--782},
  year={2023},
  publisher={IEEE}
}

@article{cai2025query,
  title={Query-aware semantic encoder-based resource allocation in task-oriented communications},
  author={Cai, Qing and Zhou, Yiqing and Liu, Ling and Yu, Hanxiao and Wu, Yihao and Shi, Ningzhe and Shi, Jinglin},
  journal={IEEE Transactions on Mobile Computing},
  volume={24},
  number={7},
  pages={6413--6429},
  year={2025},
  publisher={IEEE}
}

@inproceedings{wu2026cooperscene,
  title={CooperScene: Multi-Modal Cooperative Autonomy Benchmark with C-V2X Communication Characterization},
  author={Wu, Bo and Mo, Ruoshen and Yue, Justin and Zhang, Yanyu and Nguyen, Janice and Wu, Guoyuan and Roy-Chowdhury, Amit and Barth, Matthew J and Qiu, Hang},
  booktitle={European Conference on Computer Vision},
  pages={91--111},
  year={2026},
  organization={Springer}
}

@article{liu2023libero,
  title={Libero: Benchmarking knowledge transfer for lifelong robot learning},
  author={Liu, Bo and Zhu, Yifeng and Gao, Chongkai and Feng, Yihao and Liu, Qiang and Zhu, Yuke and Stone, Peter},
  journal={Advances in Neural Information Processing Systems},
  volume={36},
  pages={44776--44791},
  year={2023}
}

@article{xu2022cobevt,
  title={CoBEVT: Cooperative bird's eye view semantic segmentation with sparse transformers},
  author={Xu, Runsheng and Tu, Zhengzhong and Xiang, Hao and Shao, Wei and Zhou, Bolei and Ma, Jiaqi},
  journal={arXiv preprint arXiv:2207.02202},
  year={2022}
}

@article{shi2022motion,
  title={Motion transformer with global intention localization and local movement refinement},
  author={Shi, Shaoshuai and Jiang, Li and Dai, Dengxin and Schiele, Bernt},
  journal={Advances in Neural Information Processing Systems},
  volume={35},
  pages={6531--6543},
  year={2022}
}

@misc{shukor2025smolvla,
      title={SmolVLA: A Vision-Language-Action Model for Affordable and Efficient Robotics}, 
      author={Mustafa Shukor and Dana Aubakirova and Francesco Capuano and Pepijn Kooijmans and Steven Palma and Adil Zouitine and Michel Aractingi and Caroline Pascal and Martino Russi and Andres Marafioti and Simon Alibert and Matthieu Cord and Thomas Wolf and Remi Cadene},
      year={2025},
      eprint={2506.01844},
      archivePrefix={arXiv},
      primaryClass={cs.LG},
      url={https://arxiv.org/abs/2506.01844}, 
}

@inproceedings{xu2022opv2v,
  title={Opv2v: An open benchmark dataset and fusion pipeline for perception with vehicle-to-vehicle communication},
  author={Xu, Runsheng and Xiang, Hao and Xia, Xin and Han, Xu and Li, Jinlong and Ma, Jiaqi},
  booktitle={2022 International Conference on Robotics and Automation (ICRA)},
  pages={2583--2589},
  year={2022},
  organization={IEEE}
}

\appendix

\section{Scheduling Algorithm}
\label{alg:scheduling}

\begin{algorithm}[h]
\caption{Greedy Deadline-Feasible Semantic Allocation}
\label{alg:scheduling}
\small
\begin{algorithmic}[1]
\Require Pending messages $\mathcal{Q}(t)$; utility--cost profiles
$\{U_i(K), B_i(K)\}$; deadlines $\{d_i\}$; link rate $R(t)$
\Ensure Transmission sizes $\{K_i\}$ and transmission order $\pi$

\State $K_i \gets 0$, $\forall i\in\mathcal{Q}(t)$

\Repeat
    \State $\mathcal{F} \gets \emptyset$
    \ForAll{$i\in\mathcal{Q}(t)$ such that $K_i < N_i$}
        \State $k \gets K_i+1$
        \State $\rho_{i,k} \gets
        \dfrac{U_i(k)-U_i(k-1)}
        {B_i(k)-B_i(k-1)}$
        \State $\mathbf{K}' \gets \mathbf{K}$;
        $K'_i \gets K_i+1$
        \If{\Call{DeadlineFeasible}{$\mathbf{K}',t$}}
            \State $\mathcal{F} \gets
            \mathcal{F}\cup\{(i,\rho_{i,k})\}$
        \EndIf
    \EndFor

    \If{$\mathcal{F}=\emptyset$}
        \State \textbf{break}
    \EndIf

    \State $i^\star \gets
    \arg\max_{(i,\rho)\in\mathcal{F}}\rho$
    \State $K_{i^\star}\gets K_{i^\star}+1$
\Until{no feasible token increment remains}

\State $\mathcal{S}\gets
\{i\in\mathcal{Q}(t)\mid K_i>0\}$
\State $\pi\gets$ messages in $\mathcal{S}$ sorted by increasing deadline
\State \Return $\{K_i\},\pi$

\Function{DeadlineFeasible}{$\mathbf{K},t$}
    \State $\mathcal{S}\gets
    \{i\in\mathcal{Q}(t)\mid K_i>0\}$
    \State $(\pi_1,\ldots,\pi_m)\gets$
    messages in $\mathcal{S}$ sorted by increasing deadline
    \For{$q=1,\ldots,m$}
        \State $C_{\pi_q}\gets
        t+\displaystyle\sum_{\ell=1}^{q}
        \frac{8B_{\pi_\ell}(K_{\pi_\ell})}{R(t)}$
        \If{$C_{\pi_q}>d_{\pi_q}$}
            \State \Return \textbf{False}
        \EndIf
    \EndFor
    \State \Return \textbf{True}
\EndFunction
\end{algorithmic}
\end{algorithm}

We provide the detailed scheduling procedure used by \sysname in Algorithm~\ref{alg:scheduling}. We define $U_i(0)=B_i(0)=0$ for all messages $i$.
Our design separates semantic prioritization from deadline feasibility. At each iteration, it considers one additional token for every pending message and admits only increments for which all currently selected messages remain deadline-feasible under EDF ordering. Among these feasible increments, it selects the one with the largest marginal semantic gain per communication cost. The procedure terminates when no additional token can be admitted without causing a deadline violation.

\section{Detailed Settings}
\label{sec:settings}

\begin{table*}[!tp]
    \centering
    \small
    \setlength{\tabcolsep}{4pt}
    \renewcommand{\arraystretch}{1.15}
    \begin{tabular*}{\textwidth}{@{\extracolsep{\fill}}llrrl@{}}
    \toprule
    \multicolumn{5}{l}{\textbf{(a) Tasks and timing}} \\
    \addlinespace[3pt]
    Benchmark & Task & Test workload & Rate (Hz) & Deadline (ms) \\
    \midrule
    \multirow{2}{*}{CooperScene}
    & 3D detection & 3,936 jobs & 10 & 100 \\
    & Trajectory prediction (5 s) & 18,693 targets & 5 & 200 \\
    \addlinespace[3pt]
    \multirow{2}{*}{OPV2V}
    & 3D detection & 528 jobs & 10 & 100 \\
    & BEV segmentation & 265 jobs & 5 & 200 \\
    \addlinespace[3pt]
    LIBERO & Object manipulation & 45 episodes & 20 & Variable \\
    \bottomrule
    \end{tabular*}

    \par\vspace{6pt}

    \begin{tabular*}{\textwidth}{@{\extracolsep{\fill}}lrlll@{}}
    \toprule
    \multicolumn{5}{l}{\textbf{(b) Bandwidth settings (Mbps)}} \\
    \addlinespace[3pt]
    Benchmark & Full-send & Fixed rates & Fractions (\%) & Varying rates \\
    \midrule
    CooperScene & 118.5 & 15, 20, 25, 30 & 12.5, 16.7, 20.8, 25 & $30 \leftrightarrow 200$ \\
    OPV2V & 153.0 & 8, 15, 31, 61 & 5, 10, 20, 40 & $8 \leftrightarrow 61$ \\
    LIBERO & 13.46 & 3.4, 5.0, 6.7, 10.1 & 25, 37.5, 50, 75 & $3.4 \leftrightarrow 20$ \\
    \bottomrule
    \end{tabular*}
    \par\smallskip
    \begin{minipage}{\textwidth}
    \footnotesize
    Full-send reference rates are estimated from validation or calibration data.
    \end{minipage}
    \caption{Detailed experiment settings.}
    \label{tab:benchmark-settings}
\end{table*}

\begin{table*}[t]
    \centering
    \small
    \setlength{\tabcolsep}{4pt}
    \renewcommand{\arraystretch}{1.12}
    \begin{tabular}{@{}lrrrr@{}}
        \toprule
        \shortstack{Bandwidth}
        & \shortstack{Change\\evaluation}
        & \shortstack{Utility--cost\\profiling}
        & \shortstack{State update\\+ scheduling}
        & \textbf{Total} \\
        \midrule
        15 Mbps & 0.351 & 0.130 & 0.333 & \textbf{0.813} \\
        20 Mbps & 0.355 & 0.130 & 0.374 & \textbf{0.859} \\
        25 Mbps & 0.353 & 0.130 & 0.390 & \textbf{0.873} \\
        30 Mbps & 0.354 & 0.130 & 0.401 & \textbf{0.884} \\
        Varying & 0.352 & 0.129 & 0.392 & \textbf{0.873} \\
        \bottomrule
    \end{tabular}
    \caption{Mean runtime per message of \sysname on
    CooperScene (ms).}
    \label{tab:scale-runtime-overhead}
\end{table*}

\paragraph{CooperScene.}
CooperScene~\cite{wu2026cooperscene} is a real-world cooperative autonomy benchmark collected using three connected vehicles and one roadside unit. It provides synchronized LiDAR, camera, and GNSS/IMU observations, together with C-V2X communication measurements. The dataset contains approximately 59,000 LiDAR frames, 53,000 image frames, and 344,000 annotated 3D bounding boxes. Its scenes cover intersections, freeway ramps, local streets, and parking lots, supporting cooperative 3D object detection and motion prediction.

\paragraph{OPV2V.}
OPV2V~\cite{xu2022opv2v} is a simulated cooperative perception benchmark generated using OpenCDA and CARLA. It contains 73 scenes across eight CARLA towns and a digital replica of Culver City, with 11,464 frames and 232,913 annotated 3D vehicle bounding boxes. Each frame includes observations from two to seven connected vehicles, each equipped with a 64-channel LiDAR, four RGB cameras providing a combined 360-degree view, and GPS/IMU sensors. The dataset supports cooperative 3D object detection, BEV semantic segmentation, tracking, and prediction.

\paragraph{LIBERO.}
LIBERO~\cite{liu2023libero} is a benchmark for knowledge transfer in language-conditioned robotic manipulation. It contains 130 tasks organized into four suites: LIBERO-Spatial, LIBERO-Object, and LIBERO-Goal, each containing 10 tasks, and LIBERO-100, containing 100 tasks. The suites vary spatial relationships, object identities, and manipulation goals to evaluate different forms of knowledge transfer. The benchmark provides language instructions, RGB observations from workspace and wrist cameras, and robot proprioceptive states.

\paragraph{Detailed experimental settings.}
Table~\ref{tab:benchmark-settings} presents the detailed experimental settings for our evaluations on these benchmarks, including the evaluated tasks, test set sizes, task rates, deadlines, and bandwidth configurations.

\section{Additional Results}
\label{app:additional-results}
\begin{figure*}[t]
    \centering
    \includegraphics[width=\textwidth]{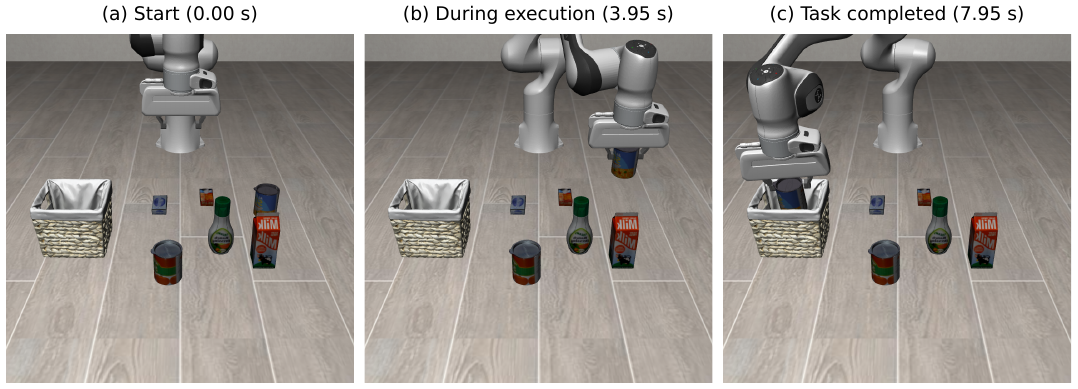}
    \caption{LIBERO task execution with SCALE at 10.1 Mbps: the robot picks up the soup can and places it in the basket. The sequence shows the start, midpoint, and successful completion of one test episode.}
    \label{fig:scene-libero}
\end{figure*}

\begin{figure*}[h]
    \centering
    \includegraphics[width=\textwidth]{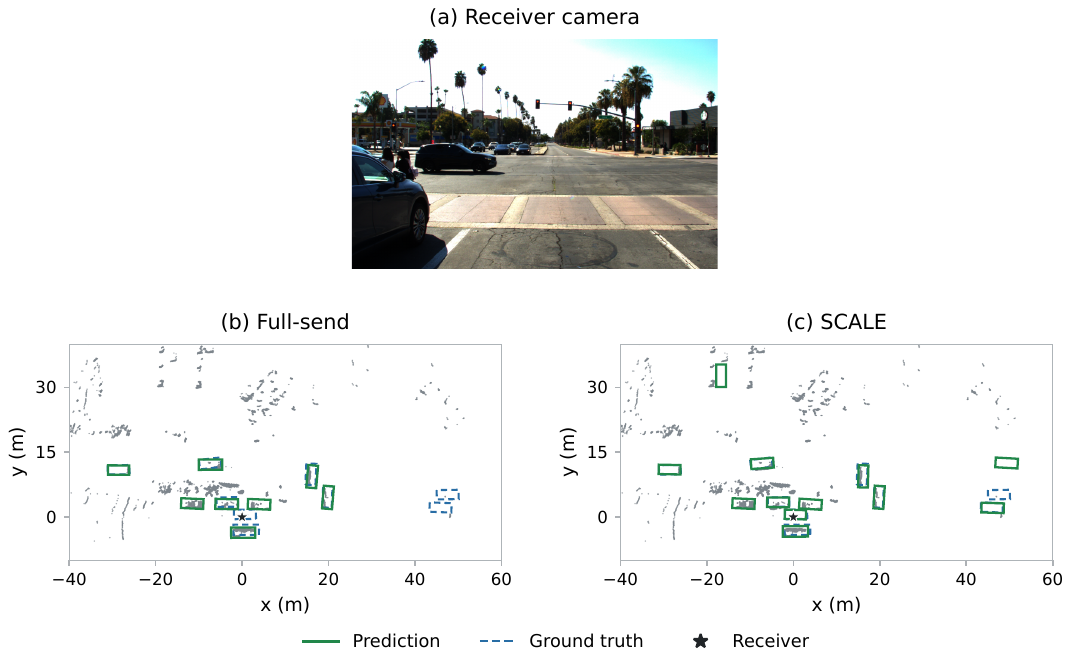}
    \caption{Vehicle detection on CooperScene at 25 Mbps. (a) Receiver camera. (b--c) Full-send and SCALE detections for the same frame, using the same CoBEVT model. Green boxes are detections; dashed blue boxes are ground truth. Both BEV panels use the same view.}
    \label{fig:scene-cooperscene}
\end{figure*}

\begin{figure*}[!tp]
    \centering
    \includegraphics[width=\textwidth]{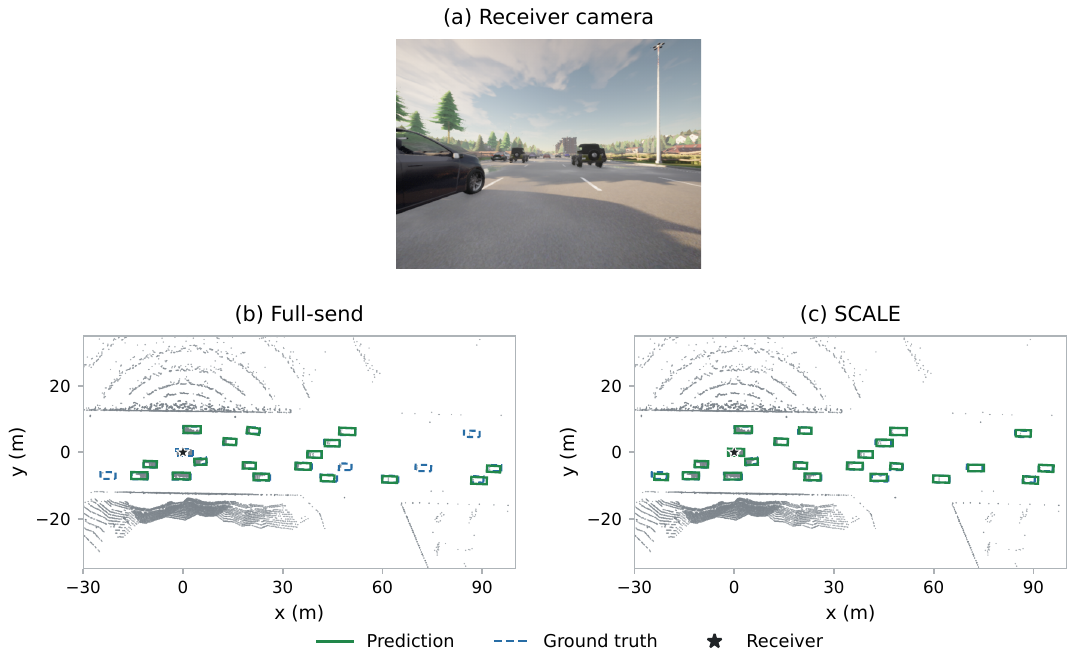}
    \caption{Vehicle detection on OPV2V at 31 Mbps. (a) Receiver camera. (b--c) Full-send and SCALE detections for the same frame, using the same CoBEVT model. Green boxes are detections; dashed blue boxes are ground truth. Both BEV panels use the same view.}
    \label{fig:scene-opv2v}
\end{figure*}

\begin{figure*}[!tp]
    \centering
    \includegraphics[width=\textwidth]{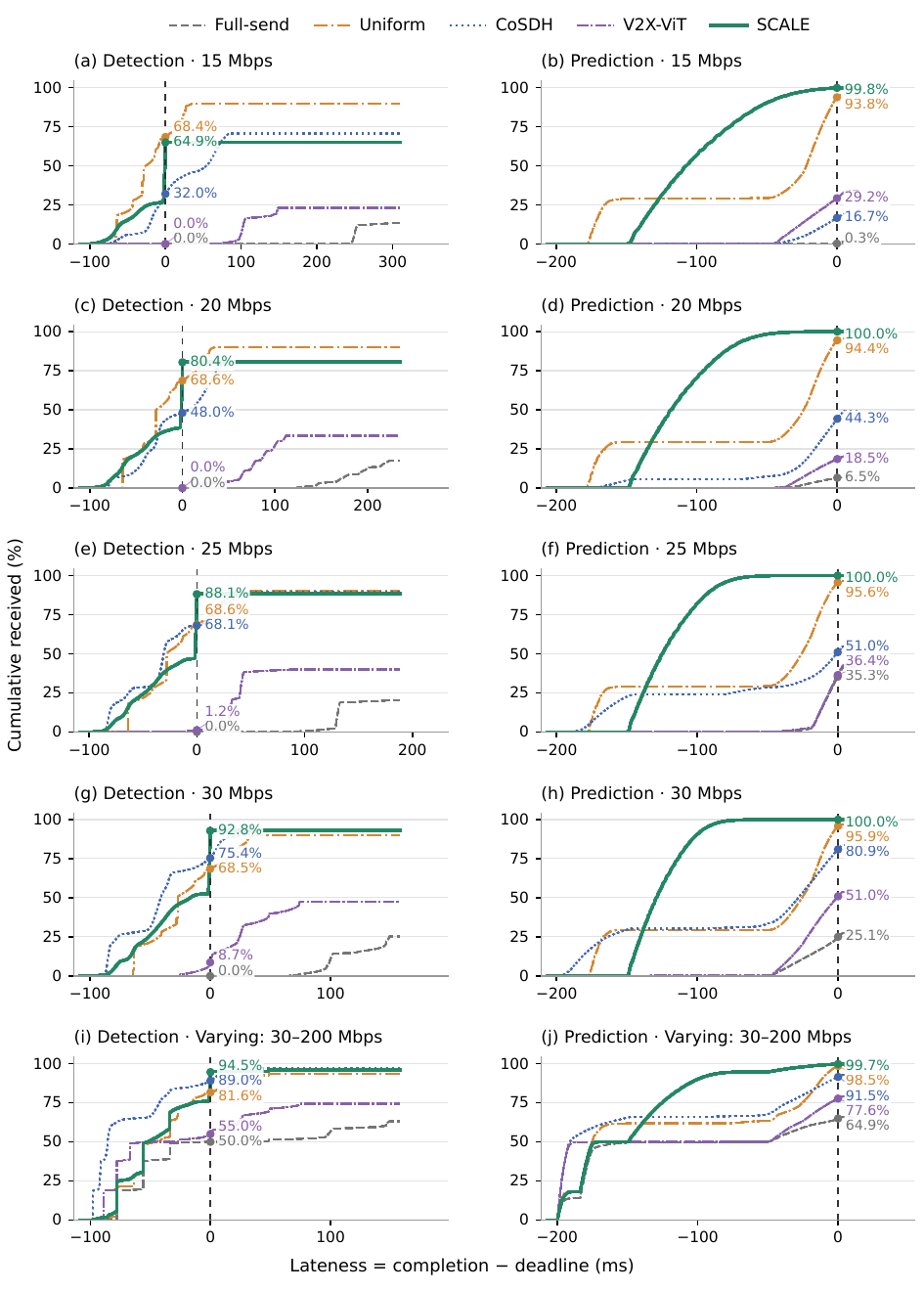}
    \caption{CooperScene message-lateness CDFs at the five selected bandwidth settings: detection (left) and prediction (right). Varying switches between 30 and 200 Mbps every 2 seconds. Zero marks each message's deadline; the CDF at zero is its method's on-time rate.}
    \label{fig:cdf-cooperscene-selected}
\end{figure*}

\begin{figure*}[!tp]
    \centering
    \includegraphics[width=\textwidth]{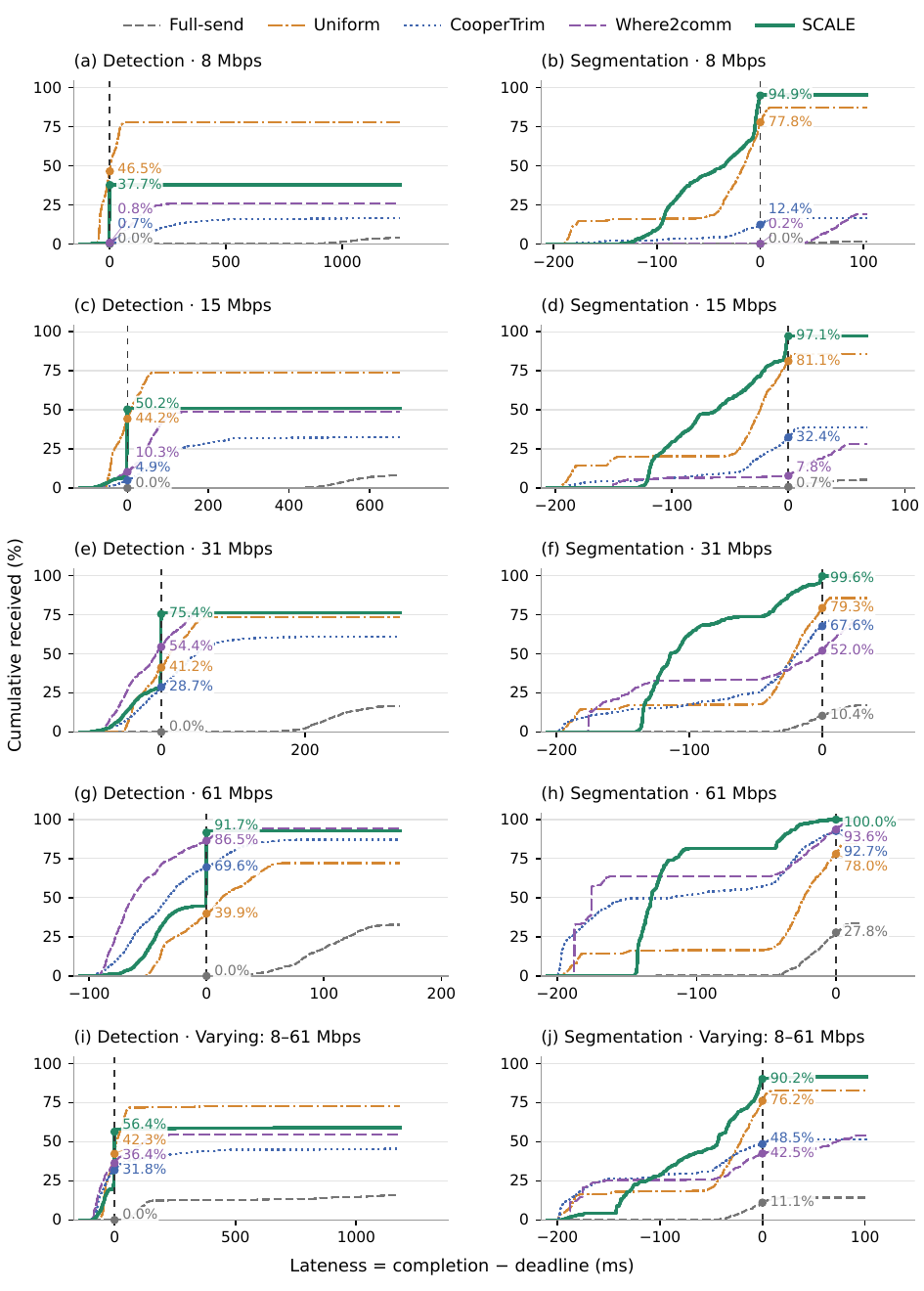}
    \caption{OPV2V message-lateness CDFs at the five selected bandwidth settings: detection (left) and segmentation (right). Varying switches between 8 and 61 Mbps every 2 seconds. Zero marks each message's deadline; the CDF at zero is its method's on-time rate.}
    \label{fig:cdf-opv2v-selected}
\end{figure*}

\begin{figure*}[!tp]
    \centering
    \includegraphics[width=\textwidth]{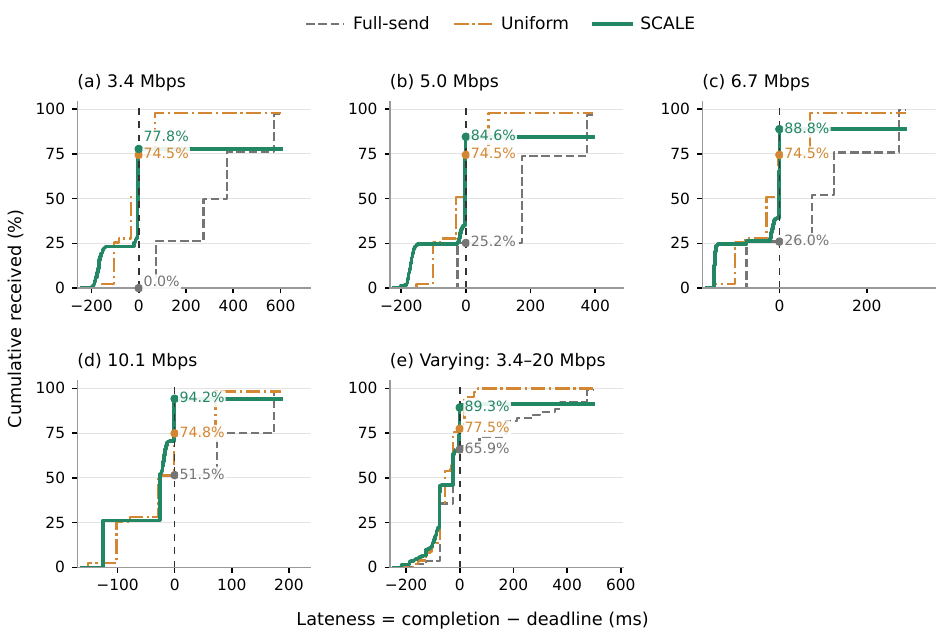}
    \caption{LIBERO camera-update lateness CDFs at the five selected bandwidth settings, using all 45 episodes per method and setting. Varying switches between 3.4 and 20 Mbps every 2 seconds. Zero marks each message's deadline; the CDF at zero is its method's on-time rate.}
    \label{fig:cdf-libero-selected}
\end{figure*}

\paragraph{Scene examples and CDFs.}
Figures~\ref{fig:scene-cooperscene} and~\ref{fig:scene-opv2v} compare the detection results of Full-send and \sysname against ground-truth annotations on CooperScene and OPV2V, respectively. Figure~\ref{fig:scene-libero} illustrates a successful LIBERO manipulation episode with \sysname. Figures~\ref{fig:cdf-cooperscene-selected}, \ref{fig:cdf-opv2v-selected}, and~\ref{fig:cdf-libero-selected} present the message lateness CDFs for CooperScene, OPV2V, and LIBERO, respectively, under fixed and varying bandwidth settings.

\paragraph{Runtime overhead.}
Table~\ref{tab:scale-runtime-overhead} reports the computational overhead of \sysname on CooperScene across five bandwidth settings. We measure runtime on an Intel Core i9-13900K using four CPU cores and four OpenBLAS threads. We report the overhead of the three design stages: semantic change evaluation, utility--cost profiling, and real-time scheduling, with the last stage also accounting for message preparation and state updates. Averaged over all generated detection and prediction messages, the total overhead remains below 1~ms per message across all settings, ranging from 0.813 to 0.884~ms. These results highlight the low computational overhead of \sysname.

\section{AI Usage Statement}
We used AI assistants to support the implementation and debugging of the experimental code described in Section 5. We also used AI assistants for grammar checking and revising the manuscript. The authors reviewed the AI-assisted code and text and tested the code. The authors take full responsibility for the experimental results and the final manuscript.

\end{document}